\documentclass[review]{elsarticle}

\usepackage{lineno,hyperref}
\modulolinenumbers[5]

\journal{ISPRS Journal of Photogrammetry and Remote Sensing}

\usepackage{wrapfig}
\newcommand{\tabitem}{~~\llap{\textbullet}~~}
\usepackage{multicol}
\usepackage{graphicx}
\usepackage{amsmath,amssymb} 
\usepackage{rotating}
\usepackage{subcaption}
\usepackage{natbib}
\usepackage{booktabs}
\usepackage{float}
\usepackage{multirow}
\newcommand{\comment}[1]{}
\usepackage[percent]{overpic}
\usepackage[dvipsnames]{xcolor}
\usepackage{array}
\newcolumntype{M}[1]{>{\centering\arraybackslash}m{#1}}

\usepackage[normalem]{ulem}
\usepackage{xr}
\usepackage{makecell}
\usepackage{arydshln}

\begin{document}

\begin{frontmatter}

\title{Water level prediction from social media images with a multi-task ranking approach}



\author{
 P. Chaudhary\textsuperscript{1}, S. D'Aronco\textsuperscript{1}, J.P. Leit\~{a}o\textsuperscript{2}, K. Schindler\textsuperscript{1}, J.D. Wegner\textsuperscript{1}
 }
 
\address{
	\textsuperscript{1 }EcoVision Lab, Photogrammetry and Remote Sensing group, ETH Z\"urich, Switzerland\\
	(priyanka.chaudhary, stefano.daronco, jan.wegner)@geod.baug.ethz.ch, schindler@ethz.ch\\
	\textsuperscript{2 }Department Urban Water Management, Eawag - Swiss Federal Institute of Aquatic Science and Technology, Switzerland\\
	(joaopaulo.leitao)@eawag.ch\\
}




\begin{abstract}
Floods are among the most frequent and catastrophic natural disasters and affect millions of people worldwide. It is important to create accurate flood maps to plan (offline) and conduct (real-time) flood mitigation and flood rescue operations.
Arguably, images collected from social media can provide useful information for that task, which would otherwise be unavailable. 
We introduce a computer vision system that estimates water depth from social media images taken during flooding events, in order to build flood maps in (near) real-time.
We propose a multi-task (deep) learning approach, where a model is trained using both a regression and a pairwise ranking loss.
Our approach is motivated by the observation that a main bottleneck for image-based flood level estimation is training data: it is difficult and requires a lot of effort to annotate uncontrolled images with the correct water depth.
We demonstrate how to efficiently learn a predictor from a small set of annotated water levels and a larger set of weaker annotations that only indicate in which of two images the water level is higher, and are much easier to obtain.
Moreover, we provide a new dataset, named \textsc{DeepFlood}, with 8145 annotated ground-level images, and show that the proposed multi-task approach can predict the water level from a single, crowd-sourced image with $\approx$11$\,$cm root mean square error. 
\end{abstract}

\begin{keyword}
Object detection, Deep learning, Image segmentation, Flood estimation, Learning to rank, Flood detection
\end{keyword}

\end{frontmatter}


\section{Introduction}\label{INTRODUCTION}


The frequency of weather-related disasters is increasing rapidly: During the period of 1995--2015, floods have accounted for $47\%$ of all weather related disasters and have affected over 2 billion people \cite{human_cost}. The number of floods has also soared up to an average of 171 floods per year between 2005-2014, compared to 127 floods per year during 1995--2004 \cite{human_cost}. Moreover, a change in the nature of these events has been observed, with an increase of flash floods, acute riverine and coastal flooding. Additionally, the progressive urbanisation has resulted in large flood run-offs~\cite{human_cost}. 

To mitigate the damage caused by such flood events and for effective disaster response and emergency plans, the rapid analysis of data collected from the affected area is essential~\cite{2019arXiv190803361B}. There are various sources from where observations can be gathered: stream gauge data~\cite{PARKES20161189,SUN20004}, remote sensing data \cite{doi:10.1002/esp.1637,TRALLI2005185} and field data collection. The field data collection approach consists of sending people to the affected areas to survey and document data after the flood event. The information collected can then be used to prepare flood-inundation maps~\cite{Musser_2016}. However, implementing this approach in real-time is expensive, labour intensive and difficult to obtain from flooded areas during, or immediately after, the flood event~\cite{zhelong}.

Data collected from stream gauges provide accurate, near real-time information of water height for the monitored locations, but gauges are sparsely distributed leading to extremely sparse observations. Stream gauges are not installed systematically along every waterway and much less away from the water streams. Due to these dispersed locations the information provided is often not sufficient to map the flooded area. In addition, stream gauges are rendered useless in cases where the water level rises beyond the limit of gauges themselves or if they are washed away during a flood event~\cite{zhelong}.

Remotely sensed satellite imagery has been widely used to monitor disaster events like floods~\cite{zhelong}. However, many satellites have comparatively long revisit cycles, which makes them less useful when information should be gathered in real time, or flood duration is relatively short, like in the case of pluvial/ localised flooding. 
In general, only the flooded area can be retrieved, whereas water depth cannot be observed directly.
Moreover, satellite sensors in the optical wavelengths are affected by the cloud cover, which is inherently frequent during flooding events. Aerial photography is another commonly data source for flood mapping, but it also dependent on weather conditions, and expensive~\cite{zhelong}.

The unprecedented global spread of low-cost sensors, especially in smartphones, together with the rise of the internet and social media, opens the possibility of community-based mapping initiatives~\cite{STARKEY2017801}. Recognition is increasing for the utility of social media when it comes to capturing real-time information during and immediately after a flood, using "citizens-as-sensors". 

In earlier work~\cite{isprs-annals-IV-2-W5-5-2019} we have presented a model to predict flood height from images gathered from social media platforms in a fully automated way using a deep learning framework. The proposed model 
performed object instance segmentation and predicted flood level whenever an instance of some specific object was detected. Although the trained model performs rather well, the effort required to build a large, pixel-accurate annotated dataset for instance segmentation of flood images is considerable. To tackle this problem, we propose in this paper a deep learning approach where we define the flood estimation as a per-image \emph{regression} problem and combine it with a \emph{ranking loss} to further reduce the labelling load. We propose to avoid the tedious, and hardly scalable, procedure of pixel-accurate object instance labelling per image by \emph{(i)} directly regressing one representative water level value per image and, more importantly, \emph{(ii)} exploiting relative ranking of the water levels in pairs of images, which is much easier to annotate.

Moving from pixel-accurate object delineation as in~\cite{isprs-annals-IV-2-W5-5-2019} to annotating only a single water depth per image comes at a price. While the regression task might, in principle, be easier than detailed object detection and segmentation, the supervision signal for a machine learning system is much weaker (e.g., we no longer tell the system to turn its attention to certain types of objects that reoccur with similar metric height).
Furthermore, even in the presence of known objects it is often hard for a human operator to determine the water depth in individual images on an absolute scale.
On the contrary it is a much simpler task to rank images via pairwise comparisons. People can, with no or little training, quickly decide which of two images shows a higher water level.
In this way it becomes feasible to outsource the labelling effort to large groups of untrained annotators, for instance through an online tool. Using ranking as a complementary task can be seen as a variant of \emph{weak supervision}, or alternatively the ranking information can be interpreted as a \emph{regulariser} for the otherwise data-limited regression task.
The idea is that a large volume of weaker ranking labels should be able to largely compensate for the small amount of strong water depth labels, and lead to better regression performance. 
%
%
We make three contributions in this paper: 

\emph{(i)} We propose a deep learning approach that learns to estimate water level from social media images by combining water level regression with a relative ranking of image pairs. The water level regression part is fully supervised while pairwise image ranking adds a weak supervision signal to improve overall accuracy. The general idea is that the fully supervised signal (i.e., water level regression) from a small, expensive label set is supported by a closely related, weak supervision signal (i.e., pairwise water level ranking), where collecting large amounts of labels is cheap.
%

\emph{(ii)} We introduce a new, large-scale dataset \textsc{DeepFlood} with  $>$8000 images. \textsc{DeepFlood} is comprised of two sub-datasets called DF-OBJ and DF-IMG which we use for our regression and ranking sub-tasks respectively (Sec.~\ref{sec:DATASETS}). We make all data available on request via email to one of the authors of this paper.

\emph{(iii)} We experimentally investigate the trade-off between an object-driven approach with pixel-accurate segmentation labels, versus a 
regression of the water level with (or without) support from weak pairwise rankings (Sec.~\ref{sec:EXPERIMENTS}).

\section{Related Work}\label{sec:RELATED WORK}

Using ground-level images for water depth estimation is still a relatively new idea about which only little literature exists. In contrast, the use of social media text has received more attention. Also learning from rank order is a well-known concept in machine learning, but has only recently been adopted to for deep learning. In the following, we review those works that are related closest to ours.
\subsection{Flood estimation from images and social media}\label{sec:Flood estimation from images}

Over the past few years there has been an increased interest to use data from social media for flood detection, mapping and estimation. These data have the advantage of being available in real-time, while at the same time being inexpensive to collect.

Wang \textit{et al.}~\cite{WANG2018139} propose to use social media and crowd-sourcing data to complement traditional remote sensing data and witness reports. In their study, they use Twitter and the MyCoast crowd-sourcing platform to collect data. MyCoast is an app that has been used by a number of US environmental agencies, since 2013, to collect "citizen science" data about various coastal hazards and incidents~\cite{WANG2018139}. For accurate location mapping they use Stanford's Named Entity Recognition (NER)~\cite{Finkel:2005:INI:1219840.1219885} tool to extract location data within the tweet text. Their approach for extracting flood depth information is based only on regular expression patterns in text, while the images are used only to detect the presence of flooding.


Starkley \textit{et al.}~\cite{STARKEY2017801} demonstrate the importance of community-based observations, also known as "citizen science". The observations used in that project were in many cases either photographs or videos. It is shown that community-based data are valuable for local flash flood events. Quantitative flood metrics are extracted manually. 


Fohringer \textit{et al.}~\cite{nhess-15-2725-2015} propose a methodology that leverages social media content to support rapid inundation mapping, including inundation extent and water depth. They stress that with this procedure information is readily available especially in densely populated, urban areas. This is important, since alternative information sources like remote sensing do not perform  well there. A main limitation is that, also in that system, the social media content is only retrieved automatically, but manually assessed for relevance and visually inspected by experts to derive inundation depth.

Other works, such as Aulov \textit{et al.}~\cite{Aulov2014AsonMapsAP}, Smith \textit{et al.}~\cite{doi:10.1111/jfr3.12154}, Li \textit{et al.}~\cite{zhelong},  also suggest to gather information from social media platforms that is useful for creating flood maps. The method proposed by Aulov \textit{et al.}, in particular, is able to determine regions free from flooding and regions which were flooded, together with a rough estimate of the flood depths, by manually inspecting street photos. In~\cite{doi:10.1111/jfr3.12154}, the authors present a real-time modelling framework to identify areas likely to have been flooded, exclusively information from social media platforms. They validate their results with data from Twitter during two 2012 flood events. Li \textit{et al.}~\cite{zhelong} instead use geo-referenced social media texts and combine them with a digital elevation model to generate a flood map.
Moreover,~\textit{The 2019 Multimedia Satellite Task: Emergency Response for Flooding Events}~\cite{flood_event} is one track of an online challenge offered by MediaEval. Sub-task \textit{Multimodal Flood Level Estimation from News} asks participants to build a binary classifier that predicts if an image contains at least one person standing with water above the knee. In contrast to our work, participants have access to additional text features from news articles and also satellite imagery~\cite{flood_event}. \cite{Quan2020} has achieved first rank in this task and propose the idea of matching the water level with
human pose to determine the level of severity of flooding.

Perhaps the closest work to ours in terms of water-level estimation is~\cite{isprs-archives-XLII-2-543-2018}. The authors propose to use smartphones and other (fixed) embedded system cameras to estimate water depth via explicit exterior orientation and detection of the water level. The method achieves accuracy levels on the centimeter-level, but heavily relies on a high-accuracy digital surface model of the scene. In our work, we do not demand any additional information other than the flood images, as it is often not available (at least not with the required accuracy).

To the best of our knowledge, our work is the first to propose a fully automated water level depth estimator from only single ground level images. The present paper extends the preliminary~\cite{isprs-annals-IV-2-W5-5-2019} to avoid extensive, pixel-accurate data annotation.

\subsection{Learning-to-rank and weak supervision}\label{sec:learning_to_rank}

Different ways have been developed to learn from relative ordering or ranking information, including pointwise~\cite{NIPS2001_2023} and list-based~\cite{Wang_2019_CVPR} approaches. 
The most widely used principle relies on pairwise ranking, i.e., one examines pairs of items and searches for a predictor whose outputs for the members of every such pair. Typical applications of learning-to-rank are in information retrieval~\cite{SALTON1988513}, natural language processing~\cite{10.5555/1214993, 10.5555/176313.176316, 10.5555/92858.92860} and data mining~\cite{10.5555/257938.257942, 10.5555/1972514}. 

There are only few works that implement learning-to-rank in the context of state-of-the-art image analysis with deep (convolutional) networks. Doughty \textit{et al.} \cite{Doughty_2018_CVPR} use a pairwise deep ranking model to rank the skill of a person performing some task (like drawing or surgery) based on videos. Their approach employs both spatial and temporal streams, in combination with a loss function designed to discriminate between comparable and different skill levels. Further applications of learning-to-rank include image quality assessment~\cite{Liu_2017_ICCV}, age estimation~\cite{Chen_2017_CVPR}, and fine-grained quantification of image similarity~\cite{6909576}. 

Most related to our work is a recently method for crowd counting in images~\cite{Liu_2018_CVPR}. The authors exploit the idea that for any image window, cropping a sub-window will result in a new window with an equal or smaller number of people. The corresponding pairs yield a self-supervised ranking objective, such that only little direct supervision with person counts is needed. As in our method, the ranking  is not a goal in itself, but serves as an auxiliary task that improves the performance of a regressor, e.g., crowd counting in~\cite{Liu_2018_CVPR}. %
In that view ranking can be interpreted as a form of weak supervision~\cite{weak_survey} that is cheaper to obtain than the "strong" supervision with ground truth regression targets, and regularises the model such that it generalises better, in spite of a small amount of "strong" training labels.
Weak supervision has already proved effective for other computer vision tasks, such as segmentation~\cite{weak_segmentation}. Our work is the first to employ deep, pairwise ranking as supervision for flood level estimation.

\section{Methodology}\label{sec:METHODOLOGY}



We formalise flood level estimation as a regression problem from raw images to a scalar depth value. Consequently, the supervision needed is one depth value per training image, which should be representative for the water depth in the depicted scene.
Compared to our earlier work that was based on explicit object (instance) segmentation~\cite{isprs-annals-IV-2-W5-5-2019} this avoids laborious pixel-accurate instance labelling. In the experiments, we show that, with the same number of training images, the naive regression approach performs worse than the object-driven one -- presumably because of the much lower information content per image of the supervision signal.
One solution could of course be to annotate a bigger training set of images with associated (scalar, per-image) flood depth values.
This indeed works, but is still hard to scale up, because annotating large amounts of images consistently with an absolute water level is hard.
Instead, we explore the possibility to add a weaker supervision signal that is easy to annotate and can readily be crowd-sourced, namely \emph{relative} pairwise flood depth. I.e., the annotator is presented with two images and has to determine whether the second one has higher or lower level than the first, which is much easier to do than estimating the absolute water depth.

\begin{figure*}[htbp!]
\includegraphics[width=1\textwidth]{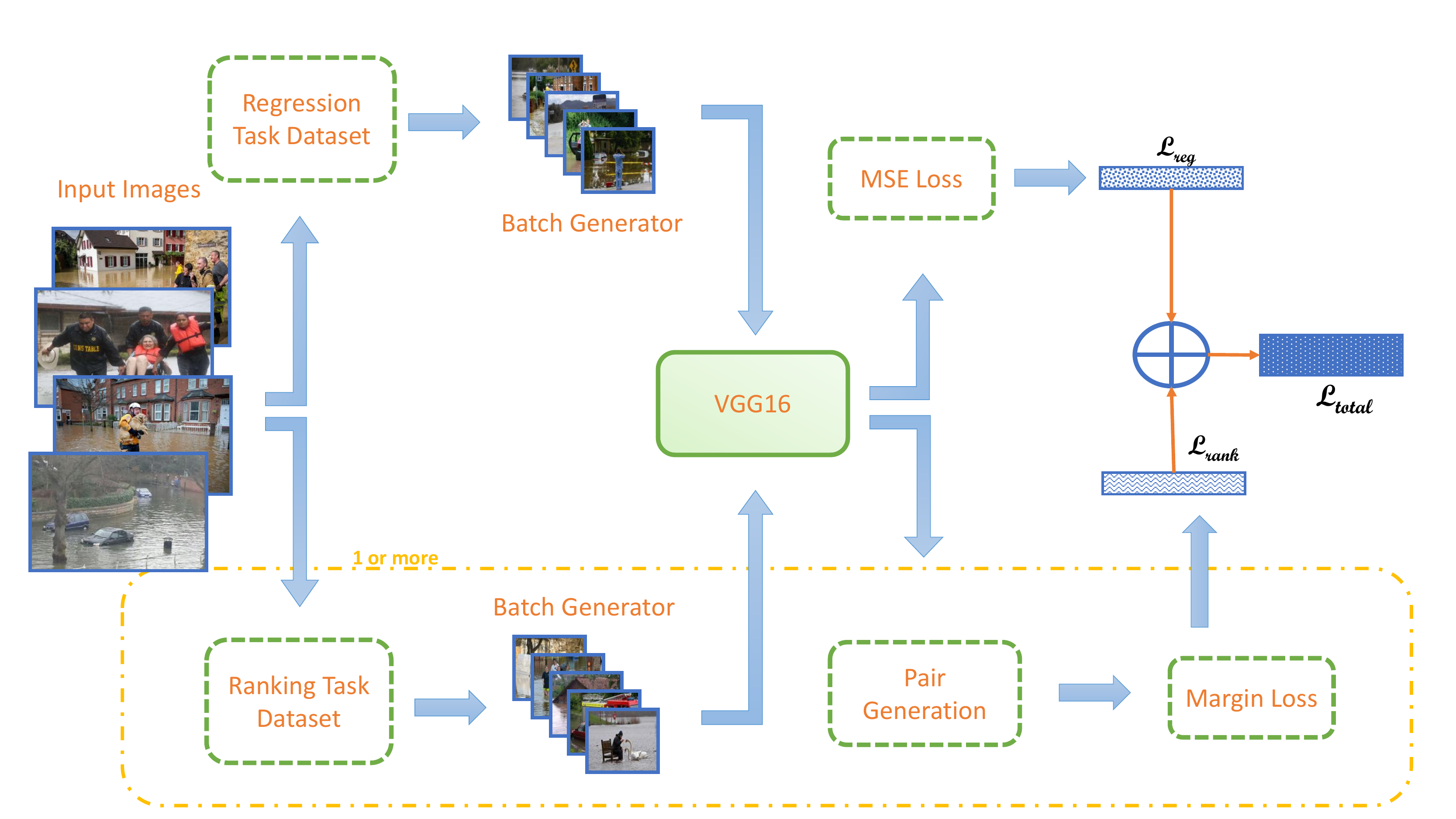}\par
\caption{The architecture of our Multi-task learning with ranking loss method. In the figure, MSE loss refers to mean squared error. $\mathcal{L}_{total}$ is the total loss function for the model. $\mathcal{L}_{reg}$ refers to the regression loss and $\mathcal{L}_{rank}$ is the ranking loss.}
\label{fig:arch}
\end{figure*}

We design a deep learning approach that combines global per-image regression and relative, pairwise ranking. The overall architecture of the proposed method is shown in Fig.~\ref{fig:arch}. The backbone of our network architecture consists of a VGG16~\cite{Simonyan15} network pre-trained on the ImageNet~\cite{5206848} dataset, but any standard network architecture could be used here. We replace the final layers of the network to predict a single scalar water depth.
Because the method does absolute water level estimation per image as well as relative ranking simultaneously, we feed two separate training sets to the model. The first part (Fig.~\ref{fig:arch} top left) has a known absolute flood water level for each image -- that is still necessary, since one cannot anchor the absolute offset and scale with only relative measurements. The second part (Fig.~\ref{fig:arch} bottom left) only knowns the ordering relation for each pair of imagess. 
Images from the regression training set are fed to the network in conventional mini-batches. For those images, we use a standard mean squared error regression loss to train the network parameters (through back-propagation).

For the images that belong to the ranking training set, the procedure is slightly different. We first prepare a mini-batch of images and feed it to the network to obtain a water level prediction for each of them. For these images we cannot evaluate the regression loss, as we do not have access to the ground truth values. We can, however, assemble all possible image pairs and test whether they obey the ground truth ranking.



We jointly learn the regression sub-task and the ranking sub-task, by defining the total loss function as:
\begin{equation}\label{eq:1}
\mathcal{L}_{total} = \mathcal{L}_{reg} + \lambda \mathcal{L}_{rank}\;,
\end{equation}
with $\mathcal{L}_{reg}$ the regression loss, $\mathcal{L}_{rank}$ the ranking loss, and $\lambda$ a weighting parameter to balance the contributions of the two terms.
For $\mathcal{L}_{reg}$, we use the Mean Squared Error (MSE) function:
\begin{equation}\label{eq:2}
\mathcal{L}_{reg} = (y - y_{\text{gt}} )^2,
\end{equation}
where $y$ represents the network output and $y_{\text{gt}}$ is the ground truth value of the flood level. 
The ranking loss $\mathcal{L}_{rank}$ is computed with:
\begin{equation}\label{eq:3}
\mathcal{L}_{rank} = \max(0, - y^{\text{rank}}_{\text{gt}}\ (y_1 - y_2) ), 
\end{equation}
where $y_1$ and $y_2$ represent the network prediction for the two images in a pair, and $y^{\text{rank}}_{\text{gt}}$ represents the ground truth ranking for the pair, where $+1$ means the level in image 1 is higher, and $-1$ means the level in image 2 is higher.
From Eq.~\eqref{eq:1} it is immediately clear that the ranking loss can be interpreted as a regularisation term that avoids overfitting of the regression objective, if the amount of training data with "strong" regression labels is limited.
The weight $\lambda$ balances the regression and ranking tasks, and must be chosen large enough to afford the regularisation, but not so high that it overpowers the regression loss and harms the prediction.
We show the influence of varying $\lambda$ empirically in Sec.~\ref{sec:EXPERIMENTS}. 

At test time, the network only receives a single image and pushes it through the regression task to obtain a flood level. It takes approximately three seconds for the model to predict a water level per image. The ranking task is not used for testing.


\section{Datasets}\label{sec:DATASETS}

We built a new dataset (\textsc{DeepFlood} \footnote{To gain access to the dataset please send us an email at priyanka.chaudhary@geod.baug.ethz.ch}) that, in total, contains 8145 ground-level images with water level annotations and extends our original dataset of~\cite{isprs-annals-IV-2-W5-5-2019}. %
From that earlier work, there are 1259 images with pixel-level object annotations. Additionally, \textsc{DeepFlood} has 5395 flood images with only a single flood depth label per image.
%
Moreover, we add 1491 images from the Mapillary Vistas dataset~\cite{Neuhold_2017_ICCV}. These images have similar characteristics and scene content as our flood images. The images from \textsc{DeepFlood} dataset are required for the network to learn how scenes from non-flooded areas look like, as the images in the \textsc{DeepFlood} dataset are from various flood events and there are no non-flooded images. %
Mapillary has pixel-level instance annotations for 37 classes, we randomly pick images from the Mapillary training set that contain at least one of the objects \textbf{Person}, \textbf{Car}, \textbf{Bus}, \textbf{Bicycle} or \textbf{Building/House} that act as basis for our water-level estimation approach.


The criteria for selecting images for \textsc{DeepFlood} and our ground truth annotation strategy remain the same as described in~\cite{isprs-annals-IV-2-W5-5-2019} because we must rely on image interpretation to annotate ground truth. We can thus only assign water levels if partially submerged objects with known average height are visible in an image. Object classes were selected based on widespread availability in social media posts of floods and roughly known average heights: \textbf{person}, \textbf{car}, \textbf{bus}, \textbf{bicycle} and \textbf{house/building}~\cite{isprs-annals-IV-2-W5-5-2019}.

%



We partition \textsc{DeepFlood} into two separate sub-datasets \textsc{DF-Obj} and \textsc{DF-Img}. \textsc{DF-Obj} contains 1862 images (1259 with flooding from our previous database, 603 Mapillary images without flooding) that all have pixel-accurate object instance annotations and annotations of the flood level per object. \textsc{DF-Img} contains 6283 images (5395 with flooding, 888 without flooding) that are annotated with a single water level per image, which is zero for images without flooding. The \textsc{DF-Obj} subset makes it possible to compare to our earlier, object-driven work~\cite{isprs-annals-IV-2-W5-5-2019}, for which instance-level segmentations are required during training. A summary of our datasets is given in Tab.~\ref{tab:dataset}.

\begin{table}
\centering
\begin{tabular}{|| m{2.2cm} | m{2.2cm} | m{2.2cm} ||}
\hline\hline
 & \textsc{DF-Obj} & \textsc{DF-Img}\\ \hline \hline
\#~Images&1862 & 6283\\ \hline
Labels&\tabitem flood level per image &\tabitem flood level per image\\ 
&\tabitem flood level per object instance & \\
&\tabitem segmentation masks & \\
\hline
Images with flooding & 1259 &5395\\ \hline
Images without flooding&603  &888\\ \hline
\hline
\end{tabular}
\caption{Overview of the \textsc{DeepFlood} dataset with 8145 images in total, with its two subsets \textsc{DF-Obj} (1862 images with pixel-accurate, object instance labels and per-image labels) and \textsc{DF-Img} (6283 images with per-image labels only).}
\label{tab:dataset}
\end{table}

Note that \textsc{DF-Obj} does not only have pixel-accurate object instance labels, but also flood level annotations per instance. The bigger \textsc{DF-Img} subset has only a single water level annotation per image. 
Consequently, the expected supervision signal passed to the model during training is much stronger for \textsc{DF-Obj}. This supervision signal explicitly shows which image regions/objects to attend to, and how their class-specific appearance changes as a function of flood depth.
We note, however, that despite the lack of object annotations the regression network may to some degree have a notion of semantic objects, since its backbone is VGG16 pre-trained on ImageNet.

We apply stratified cross-validation for experiments and report average performance numbers together with standard deviations. We divide the \textsc{DF-Obj} subset into six parts such that each part contains an equal number of images for each flood level. For each fold we use four parts for training, one for validation and one for testing.
To test direct regression methods, we simply divide \textsc{DF-Img} into a training and validation part at a ratio of $80:20$, and apply the model trained with that split to each of the six test folds of \textsc{DF-Obj}.



%
%

\begin{figure}[H]
\centering
\begin{subfigure}[htbp!]{1\textwidth}
   \includegraphics[width=1\linewidth]{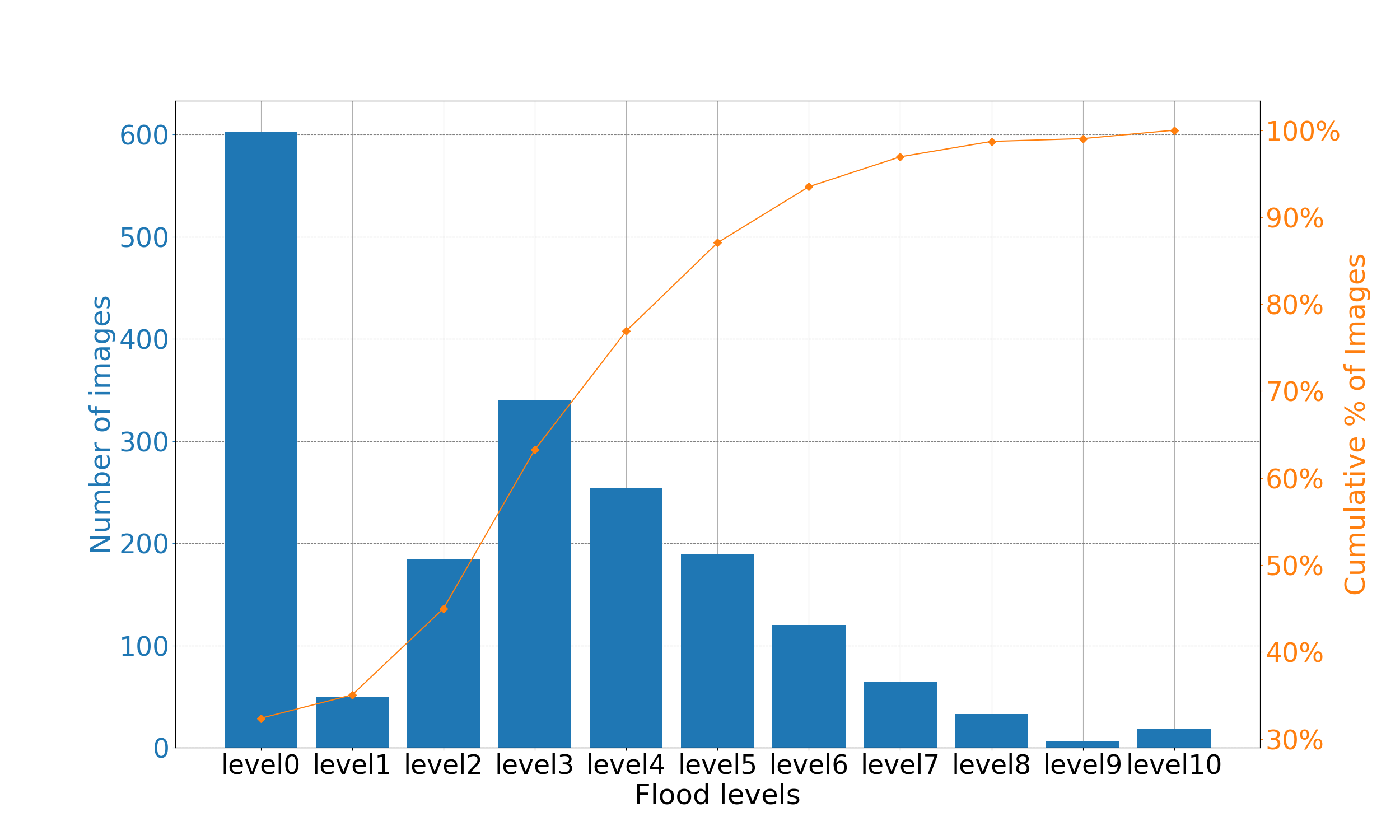}
   \caption{}
   \label{fig:Ng1} 
\end{subfigure}

\begin{subfigure}[htbp!]{1\textwidth}
   \includegraphics[width=1\linewidth]{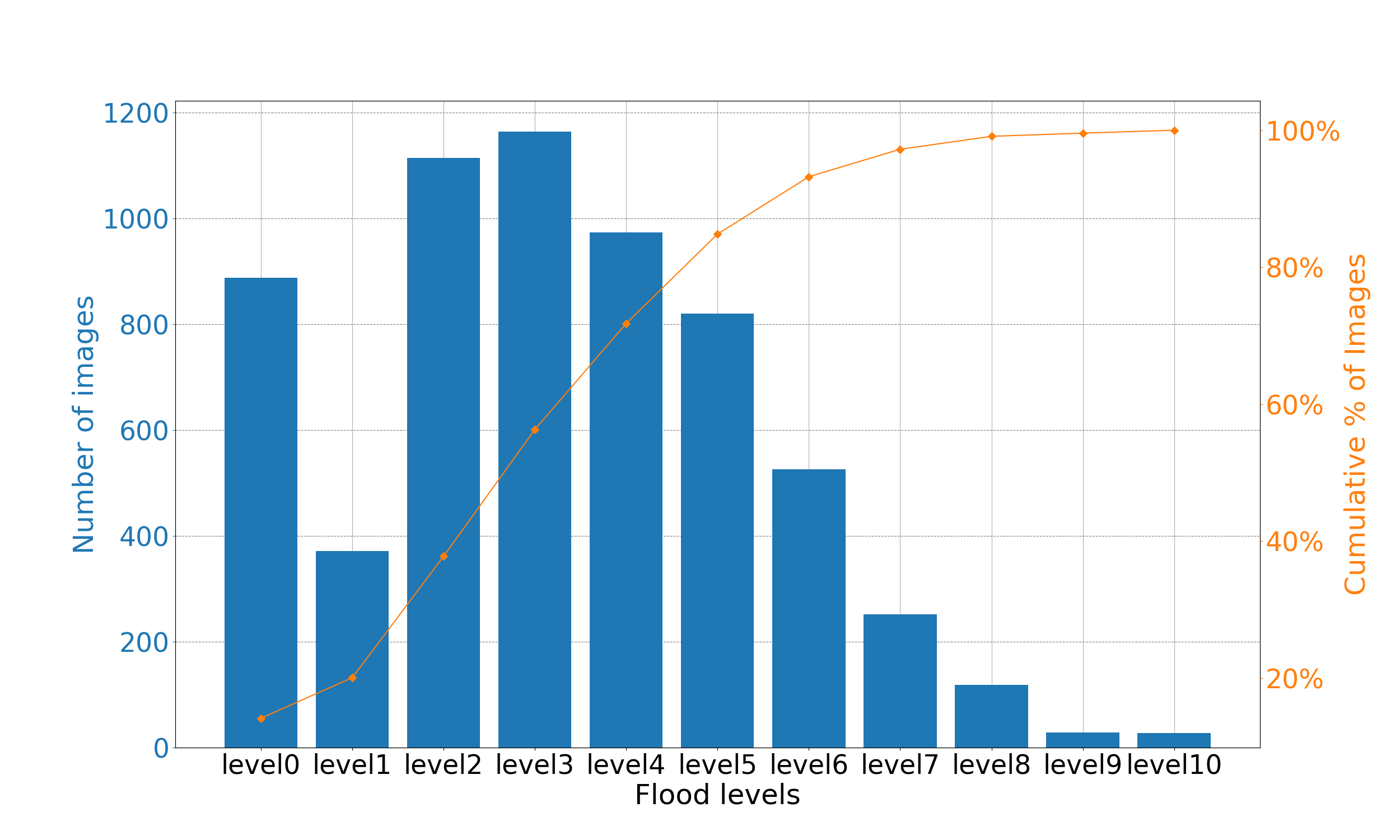}
   \caption{}
   \label{fig:img_per_level}
\end{subfigure}

\caption[Number of images per level]{Number of images per water level (including Mapillary Vistas images at level 0) for (a) the \textsc{DF-Obj} subset and for (b) the \textsc{DF-Img} subset.}
\end{figure}

We plot the amount of images per water level for both data subsets \textsc{DF-Obj} and \textsc{DF-Img} in Fig. \ref{fig:Ng1} and \ref{fig:img_per_level} respectively. Naturally, the amount of images for higher levels like $level9$ and $level10$ is much smaller compared to other levels because people are less likely to acquire images if standing in water deeper than $1.5m$. 

\section{Experiments}\label{sec:EXPERIMENTS}

We evaluate our method (\textit{Reg+Rank}) on the \textsc{DeepFlood} dataset and compare against ~\cite{isprs-annals-IV-2-W5-5-2019} (\textit{Classification}), and two baselines approaches (\textit{Regression} and \textit{Regression++}):
\begin{enumerate}
    \item \textit{Regression}: A pure regression network without additional supervision with ranked pairs, equivalent to \textit{Reg+Rank} with only the regression loss, trained on \textsc{DF-Obj}. This regression-only approach with a small training set and no pair regularisation serves as sanity check and lower performance bound.
    
    \item \textit{Regression++}: Uses the same network and loss function as \textit{Regression}, but is trained on a combination of \textsc{DF-Obj} and \textsc{DF-Img}, using absolute water levels for all training images as supervision. This corresponds to the idea case where strong supervision by regression targets is available for the entire training dataset, and serves as an upper bound for the possible performance of \textit{Reg+Rank}.

    \item \textit{Classification}: This is the object-driven approach~\cite{isprs-annals-IV-2-W5-5-2019}, where water levels are predicted via object detection and segmentation, using pixel-accurate object instance masks as supervision. Here we use use a ResNet101~\cite{He_2016_CVPR}  and Feature Pyramid Network (FPN)~\cite{Lin_2017_CVPR} as backbone and train on the \textsc{DF-Obj} subset, for which the necessary ground truth masks are available.
    
    
 
    \item \textit{Reg+Rank}: We evaluate our proposed multi-task ranking approach, which combines ranking loss and regression loss, as described in Sec.~\ref{sec:METHODOLOGY}. We train the regression loss on \textsc{DF-Obj} with the absolute water level labels per image like for \textit{Regression}. Our ranking loss is trained on the \textsc{DF-Img} data subset but, unlike \textit{Regression++}, without using absolute water levels per image. Instead, each image inside a pair of images is only labelled to have either have a lower, equal, or higher water level than the other image.  
\end{enumerate}

At inference time, only the regression task is used for prediction and performance evaluation, and it is evaluated using the root mean squared error (RMSE).

\subsection{Implementation details and parameters settings}\label{sec:evalstrat}
\begin{sidewaysfigure*}[htbp!]
\includegraphics[width=1\textwidth]{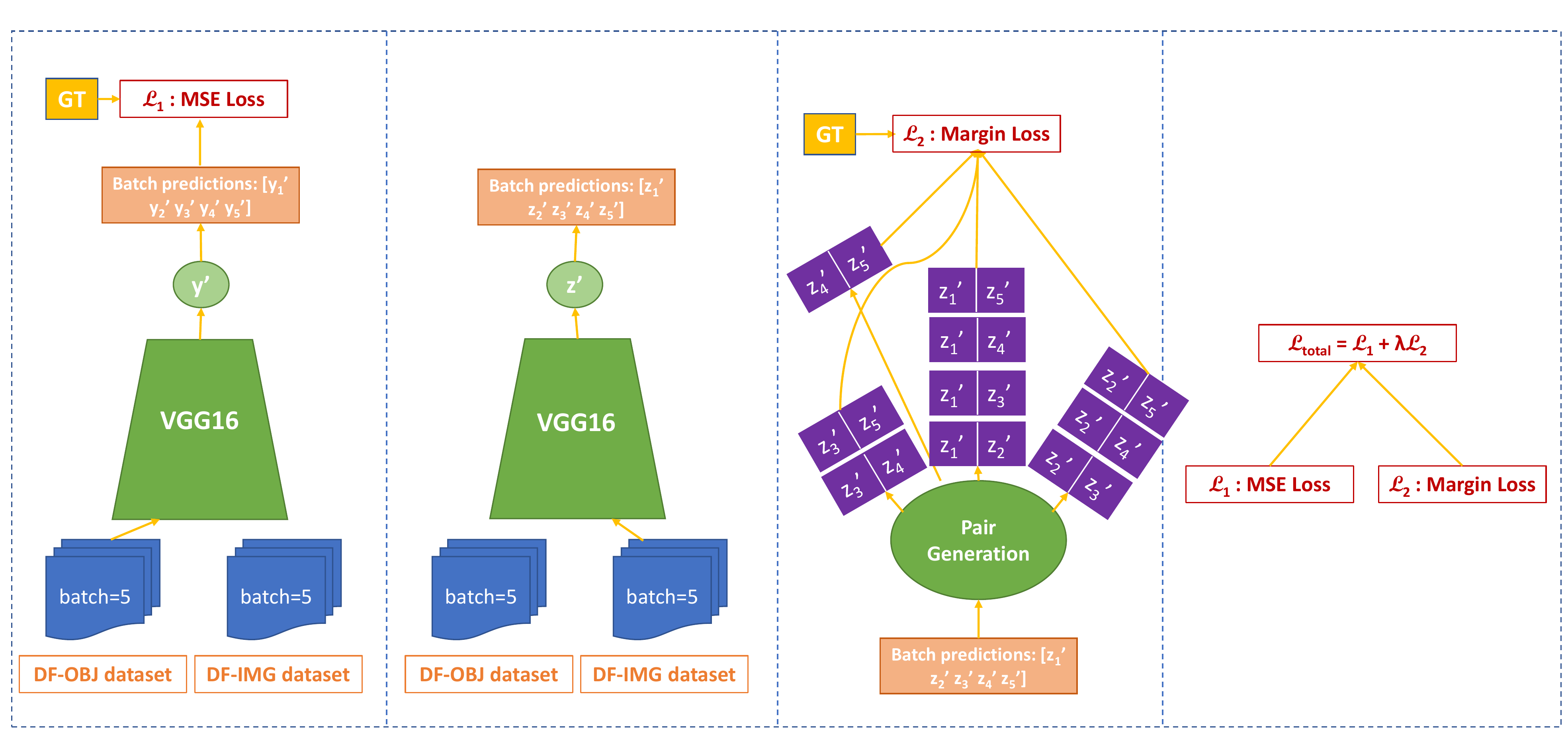}\par
\caption{Overview of the processing steps of the multi-task approach (\textit{Reg+Rank}).}
\label{fig:process}
\end{sidewaysfigure*}
We implemented all methods with the PyTorch~\cite{NEURIPS2019_9015} framework. Fig.~\ref{fig:process} provides an overview of running our approach as described in the following.

First, the images of the mini-batch generated from the \textsc{DF-Obj} dataset is passed through the VGG16 network. The batch predictions are then used with the ground truth to calculate the mean square error (MSE) for the iteration step. 
Next, we generate a mini-batch from the \textsc{DF-Img} dataset and pass it through the same VGG16 network. The newly generated predictions are then used for the image pairs generation.
We take the predictions of the mini-batch from the last step and generate distinct image pairs. Note that, for efficiency, the images need not be passed through the backbone multiple times. Rather, the pairs can be generated after feature extraction, using the single-stream Siamese network method~\cite{NIPS1993_769}: the images are first passed through the backbone in a mini-batch, then their resulting feature encodings are combined into an set of pairs (this can be viewed as a special "pair generation layer" without trainable parameters) before calculating the loss~\cite{8642842}.
Finally, the regression and ranking (margin) losses are combined.

All images were resized to $512\times512\times3~pixels$ before being fed to the model. All models were trained for $200$ epochs using the Adam optimiser~\cite{DBLP:journals/corr/KingmaB14} with an initial learning rate of $10^{-3}$. The learning rate was decreased by a factor of $10$ at epoch $150$ and $180$ for all experiments -- while Adam in theory adapts the learning rate, it empirically nevertheless makes sense to add a such a gradual schedule. We use mini-batch size $5$ for all experiments. For the ranking loss of \textit{Reg+Rank}, the $5$ images per batch are compared exhaustively, resulting in ten pairs per batch (recall, for $n$ items there are $n(n-1)/2$ distinct pairings).


We use VGG16~\cite{Simonyan15} pre-trained on ImageNet~\cite{5206848} as network backbone for \textit{Reg+Rank}, \textit{Regression}, and \textit{Regression++}, as \cite{Liu_2018_CVPR} found it to work well in the context of ranking-supported regression.
On the contrary, the object-driven classification approach is an extension of Mask R-CNN~\cite{He_2017_ICCV}, hence we use a ResNet-101-FPN backbone, as suggested by the creators of Mask R-CNN.

As usual in the transfer learning setting, we remove the top layer of VGG16 for regression and replace it, in our case with a linear layer that combines all input features into a single water level prediction.

%
%

\begin{figure}[H]
\includegraphics[width=1\textwidth]{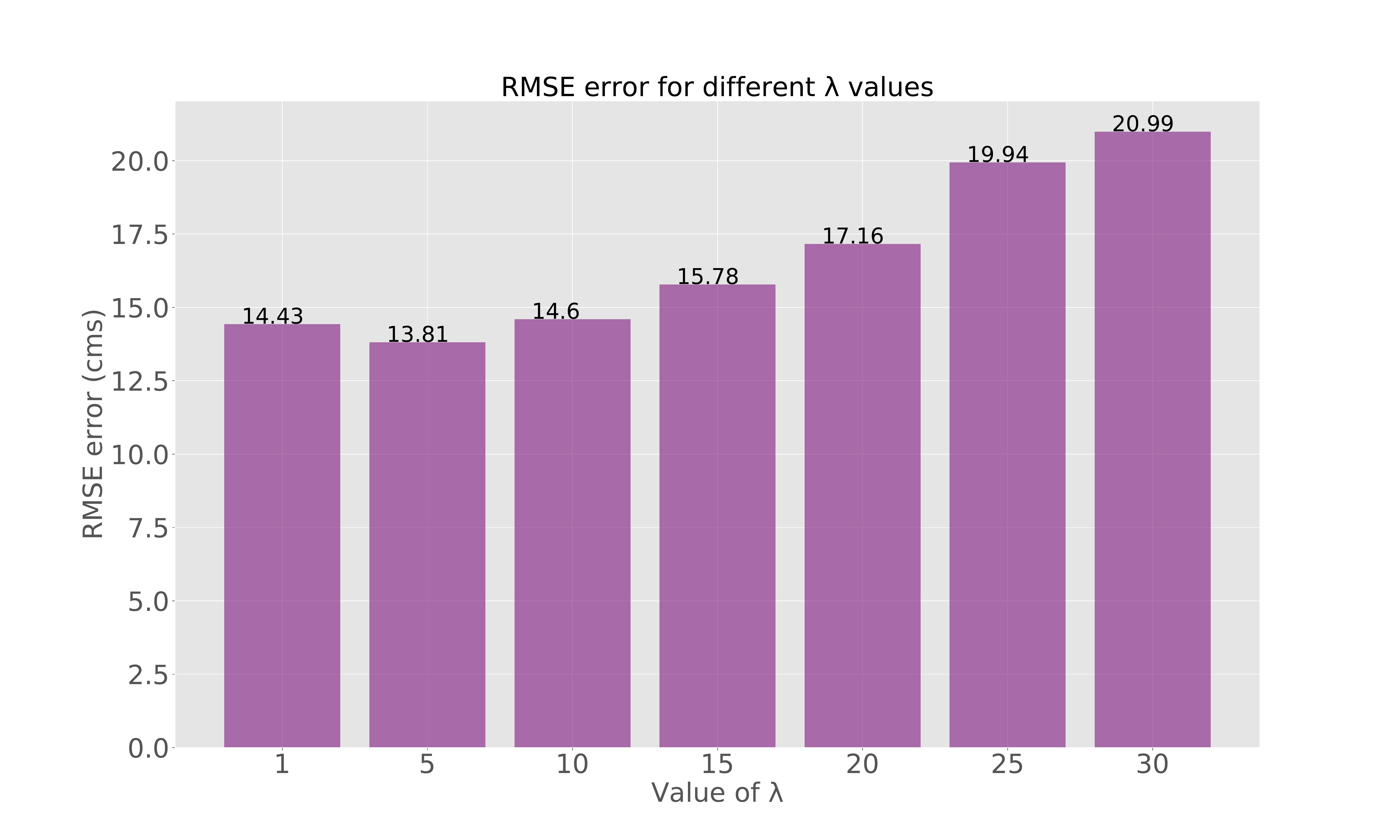}\par
\caption{Illustration of RMSE error on validation set to select best $\lambda$ value.}
\label{fig:lambda}
\end{figure}

To set the appropriate $\lambda$ parameter for \textit{Reg+Rank} (Eq.~\eqref{eq:1}), we run  experiments in which we vary $\lambda$ between $1$ and $30$ and evaluating model performance on the validation set. Figure~\ref{fig:lambda} shows the RMSE error of models with different $\lambda$ values. The lowest RMSE is achieved with $\lambda = 5$, which we use for all further experiments.

\subsection{Evaluation strategy}\label{sec:evalstrat}


Estimating an absolute water level from individual images is hard for humans. We thus pursue the strategy also used in our previous work~\cite{isprs-annals-IV-2-W5-5-2019} and look for partially submerged objects of roughly known size as "scale bars". While different objects (vehicles, bikes, etc.) are used, we discretise the depth according to the human anatomy, which provides a reasonably fine-grained set of body parts/joints that can be identified as submerged or visible.

\begin{figure*}[!htbp]
\includegraphics[width=1\textwidth]{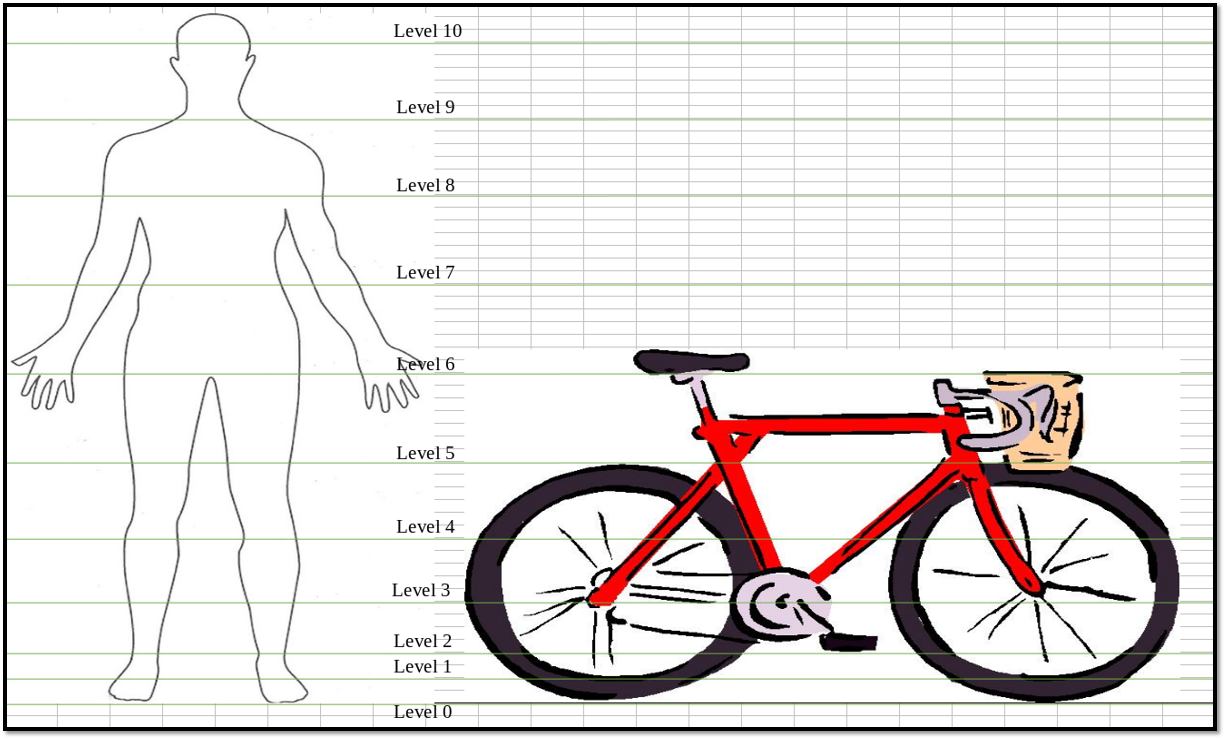}\par
\caption{Water level annotation strategy for person and bicycle}
\label{fig:annotation}
\end{figure*}

We show how flood levels are defined for humans and how it translates for an average size Bicycle in Fig.~\ref{fig:annotation}. The $11$ depth levels go from $0$, which means no water to $10$ which means a human body of average height is completely submerged in water. To convert them to metric heights we use an average human height of 170$\,$cm, see Tab.~\ref{table:level_to_cms}.
%

\begin{table}[H]
	\centering
		\begin{tabular}{||l|c|c||}\hline\hline
			Level Name& Range &Value nearest integer\\\hline\hline
			  &cm&cm\\
			 level0 & No water & 0.0 \\
			 level1 & 0.0 - 1.0 & 1.0 \\
			 level2 & 1.0 - 10.0 & 10.0 \\
			 level3 &	10.0 - 21.25	&21.0 \\
			 level4 &	21.25 - 42.5&	43.0 \\
			 level5 &	42.5 - 63.75&	64.0 \\
			 level6 &	63.75 - 85	&85.0 \\
			 level7 &	85.0 - 106.25	&106.0 \\
			 level8 &	106.25 - 127.5	&128.0 \\
			 level9 &	127.5 - 148.75	&149.0 \\
			 level10 &	148.75 - 170.0 	&170.0 \\\hline\hline
		\end{tabular}
	\caption{Definition of discrete water levels.}
\label{table:level_to_cms}
\end{table}

 In order to be able to compare the classification approach~\cite{isprs-annals-IV-2-W5-5-2019} to the regression methods, we have to map per-object flood labels to a global water depth.
 To that end, we first check if all objects are assigned \textit{level0} -- in that case we set the the global water depth to zero.
 Otherwise, we discard all objects with \textit{level0}, as even in the presence of flood water some objects may be located outside of the water (e.g., on bridges, balconies or boats) and should therefore  not contribute to the water depth estimate. The non-zero flood levels are averaged and converted to metric scale using Tab.~\ref{table:level_to_cms}.
As a measure for the deviation the predicted flood heights $\hat{y_i}$ and the ground truth $y_i$, we compute the root mean square error (RMSE) over all $n$ images, $\text{RMSE} = \sqrt{\frac{1}{n}\sum_{i=1}^{n}{\Big(\hat{y_i} -y_i)^2}}$.


\subsection{Results}\label{sec:Results}

In this section, we compare the proposed multi-task ranking approach (\textit{Reg+Rank}) with \textit{Regression}, \textit{Regression++} and \textit{Classification}~\cite{isprs-annals-IV-2-W5-5-2019}. All results are shown in Tab.~\ref{tab:results}. 
As expected, \textit{Reg+Rank} outperforms \textit{Regression} trained only on the \textsc{DF-Obj} data subset. The $\approx$22\% drop in RMSE (cm) is the benefit one gets from additional ranked pair supervision.
More interestingly, the multi-task (\textit{Reg+Rank}) approach performs almost on par with the upper bound \textit{Regression++} trained with strong supervision from the entire training data. I.e., up to a small difference of $\approx$3.5\% the ranking information can compensate for the 5$\times$ larger training set.

Since metric estimates in centimeters for water-level predictions relies on ground truth that is a function of average object sizes (i.e., it may vary slightly for each individual object instance in the images) and our manual image labelling strategy, which introduces additional uncertainties, we additionally report average RMSE (avgRMSE) across water levels (Tab.~\ref{tab:results}, two rightmost columns). Similar to the evaluation in centimeters, there is $\approx$ 21\% decrease in avgRMSE from \textit{Regression} to our \textit{Reg+Rank} approach. The avgRMSE reported in both centimeter and water-level for all experiments generally follows a similar trend. The water-level avgRMSE for \textit{Reg+Rank} is almost the same as our upper bound \textit{Regression++} experiment. It should be kept in mind though that not all water-level intervals have equal size. For example, the interval size of level2 is smaller than the interval size of level8.

\begin{table}[t]
\centering
\begin{tabular}{|| c | c | c | c | c || c | c ||}
\hline\hline
\multirow{2}{2.5cm}{\textbf{Experiments}} & \multirow{1.5}{*}{\textbf{avgRMSE}} & \multirow{1.5}{*}{\textbf{stdDev}} &
\multirow{1.5}{*}{\textcolor{black}{\textbf{avgRMSE}}} &
\multirow{1.5}{*}{\textcolor{black}{\textbf{stdDev}}} \\ 
& [cm]& [cm] & \textcolor{black}{[level]}  & \textcolor{black}{[level]} \\%
\hline\hline%
\textit{Regression}  & 14.4 & 0.45 & \textcolor{black}{0.78} & \textcolor{black}{0.01}\\

\textit{Regression++} & 10.9 & 0.85 & \textcolor{black}{0.61} & \textcolor{black}{0.05}\\ 

\textit{Classification}~\cite{isprs-annals-IV-2-W5-5-2019} & 13.6 & 0.70 & \textcolor{black}{0.80} & \textcolor{black}{0.03}\\ 

\textit{Reg+Rank} & 11.3 &0.64 & \textcolor{black}{0.62} & \textcolor{black}{0.03}\\
\hline\hline

\end{tabular}
\caption{Quantitative results of experiments on the \textsc{DeepFlood} dataset. \textit{Regression} and \textit{Classification} are evaluated using only the \textsc{DF-Obj} data subset (using per-image labels for \textit{Regression}, while \textit{Regression++} and \textit{Reg+Rank} are evaluated on both data subsets \textsc{DF-Obj} and \textsc{DF-Img} (i.e., the whole \textsc{DeepFlood} dataset) using per-image labels. We report the average root mean square error (avgRMSE) and its standard deviation (stdDev) for 5-fold cross-validation, in centimeters and level.}
\label{tab:results}
\end{table}

It was expected that \textit{Regression} performs 
worse than \textit{Regression++} and \textit{Reg+Rank}. The comparison to \emph{Classification} was less clear, but also that method performs 
worse, and only a little better than the baseline \textit{Regression} approach. I.e., the richer semantic segmentation labels and associated object knowledge bring a moderate improvement ($\approx$~6\%) over the baseline, but cannot overcome the disadvantage of the smaller \textsc{DF-Obj} training set.
As \textit{Classification} has, theoretically, the strongest supervision signal of all four approaches, we assume that it would ultimately perform as well as \textit{Regression++}, or even better, if it had access to pixel-accurate object masks for the full \textsc{DF-Img} data subset, too. %
However, it would constitute a huge effort to manually label thousands of images at that level of detail. This was a main motivation for the multi-task ranking approach \textit{Reg+Rank}, and indeed, the seemingly weaker signal via pairwise relative ranking of (thousands of) image pairs does bring a marked improvement. Large-scale collection of fast, inexpensive ranked pairs appears to be a viable alternative, especially considering how much easier it is to crowd-source to untrained workers or volunteers at scale.

\begin{figure*}[htbp!]
\includegraphics[width=1\textwidth]{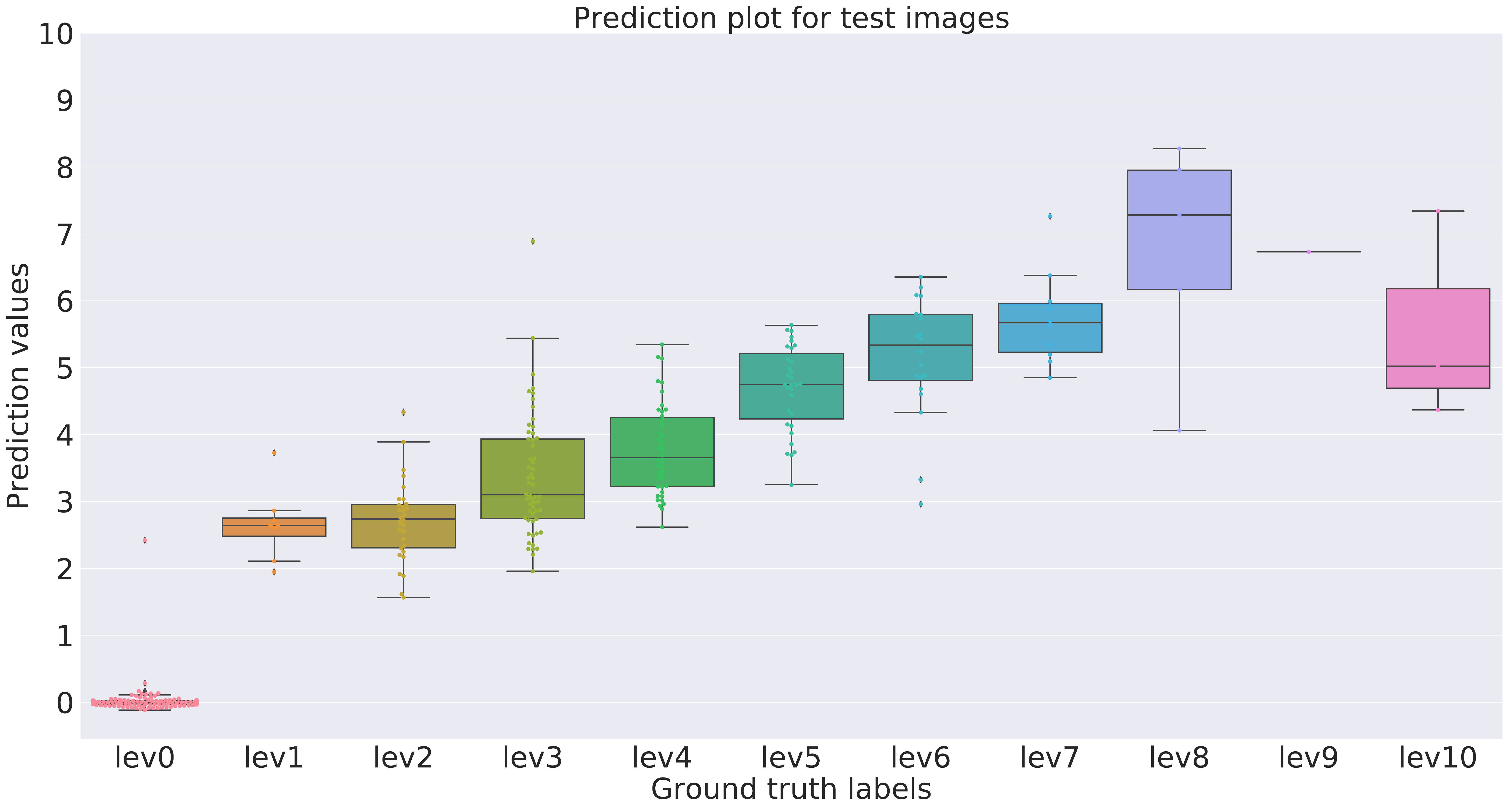}\par
\caption{Prediction plot for fold2 test images}
\label{fig:box2}
\end{figure*}

\begin{figure*}[htbp!]
\includegraphics[width=1\textwidth]{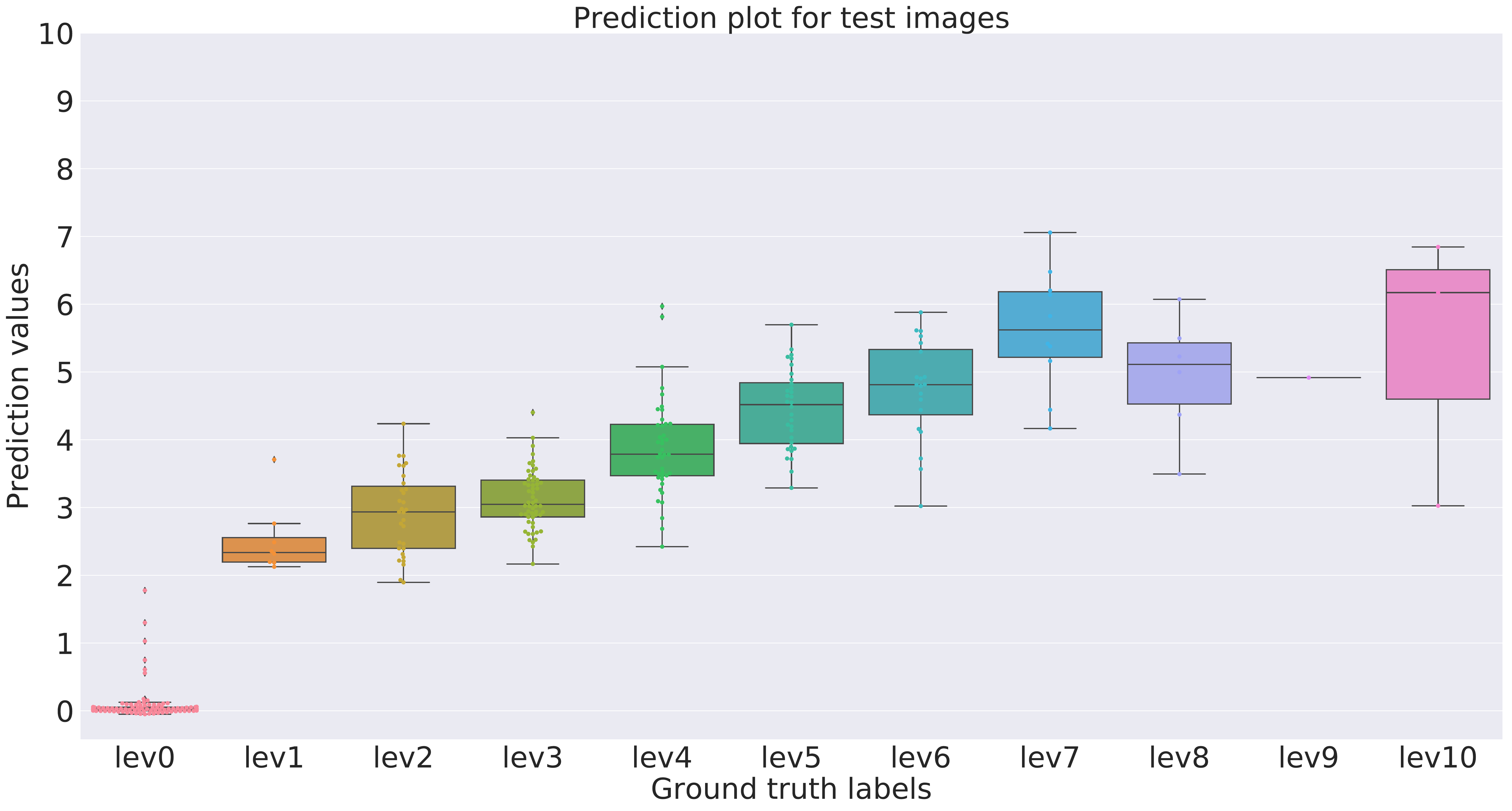}\par
\caption{Prediction plot for fold5 test images}
\label{fig:box5}
\end{figure*}

We further display the distribution of the water level predictions from our multi-task ranking approach on the cross-validation folds where \textit{Reg+Rank} performs best (fold2, Fig.~\ref{fig:box2}) and worst (fold5, Fig.~\ref{fig:box5}).
In general, \textit{Reg+Rank} tends to overestimate low water levels and underestimate very high water levels. We point out that high water levels are in general underrepresented in the data, as people are less likely to capture and upload images in such extreme circumstances. E.g., for the very high water level \textit{level9} we have only a single image in each of the two displayed folds.

\begin{figure*}[htbp]
\centering
\begin{tabular}{|c|c|c|}
\hline
\multirow{4}{*}{a} & \multirow{4}{*}[9pt]
{
\begin{overpic}
[width=0.75\textwidth, height=5.2cm]
{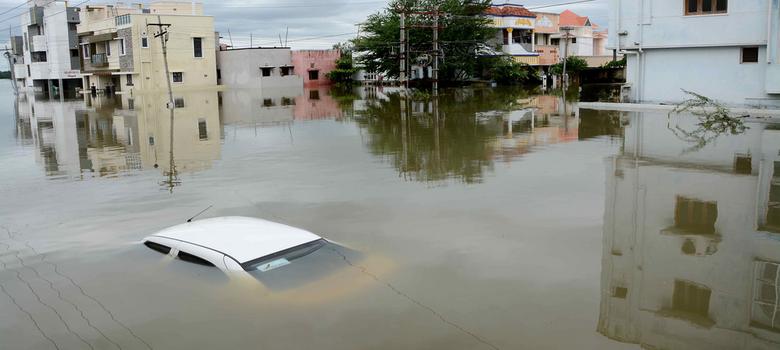}
\put(19,1)
{\color{white}\huge\textbf{Ground truth: 9}}
\end{overpic}
}
& \rule{0pt}{0.75cm}
{
\makecell
{
\dashuline{\textit{Regression}}\\
4.9}
}
\\
\cline{3-3}
& & \rule{0pt}{0.75cm}
{\makecell{\dashuline{\textit{Regression++}}\\ 5.4}}\\
\cline{3-3}
& & \rule{0pt}{0.75cm}
{\makecell{\dashuline{\textit{Classification}~\cite{isprs-annals-IV-2-W5-5-2019}}\\ 5.0}}\\
\cline{3-3}
& & \rule{0pt}{0.75cm}
{\makecell{\dashuline{\textit{Reg+Rank}}\\ 4.9}}\\
\hline
\hline

\multirow{4}{*}{b} & \multirow{4}{*}[12pt]{\begin{overpic}[width=0.75\textwidth, height=6.0cm]{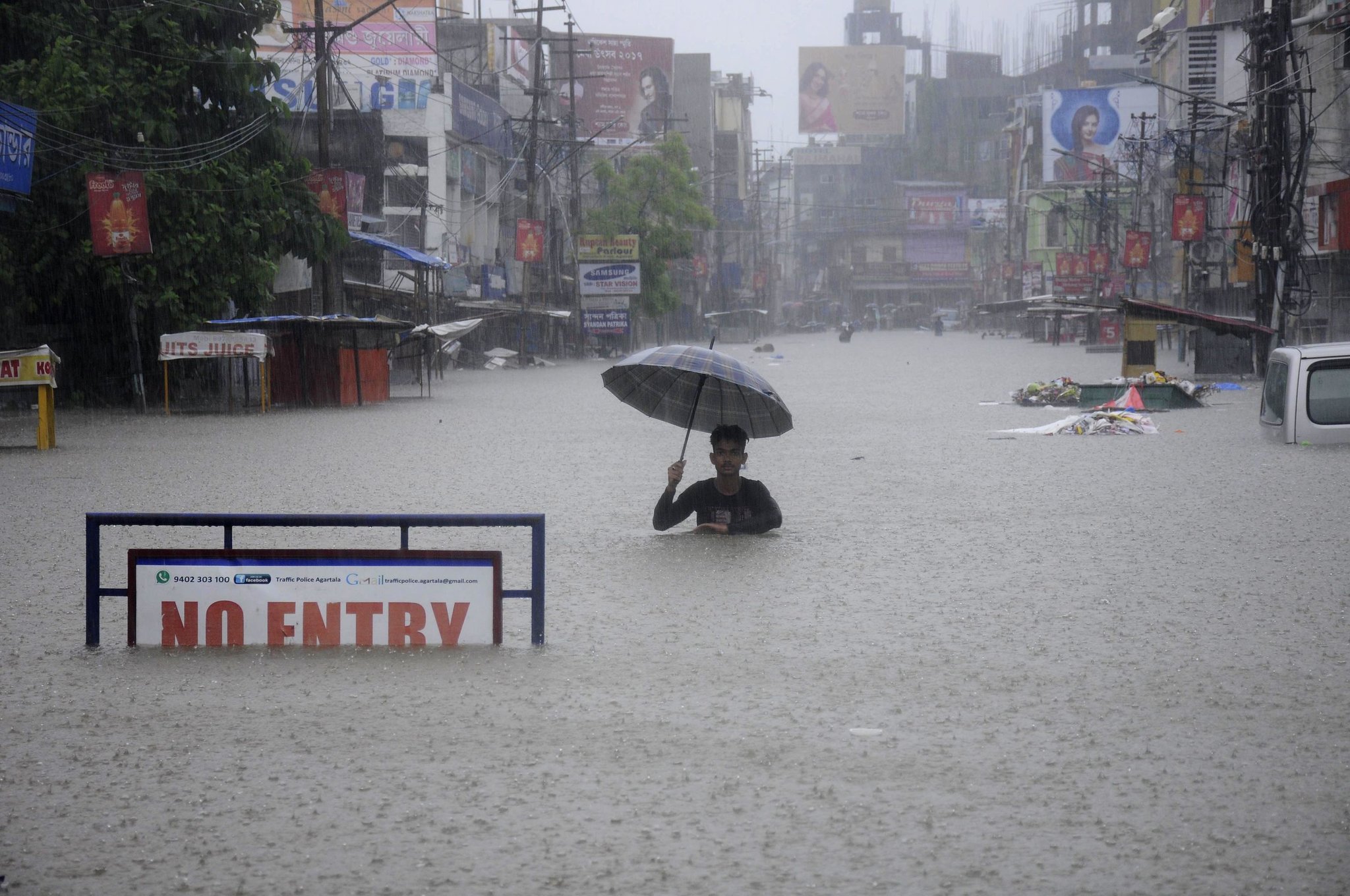}
\put(19,1){\color{white}\huge \textbf{Ground truth: 7}}
\end{overpic}}
& \rule{0pt}{1cm}
{\makecell{\dashuline{\textit{Regression}}\\ 5.3}}\\
\cline{3-3}
& & \rule{0pt}{1cm}
{\makecell{\dashuline{\textit{Regression++}}\\ 5.5}}\\
\cline{3-3}
& & \rule{0pt}{1cm}
{\makecell{\dashuline{\textit{Classification}~\cite{isprs-annals-IV-2-W5-5-2019}}\\ 6.0}}\\
\cline{3-3}
& & \rule{0pt}{1cm}
{\makecell{\dashuline{\textit{Reg+Rank}}\\ 5.8}}\\
\hline
\hline

\multirow{4}{*}{c} & \multirow{4}{*}[12pt]{\begin{overpic}[width=0.75\textwidth, height=6.0cm]{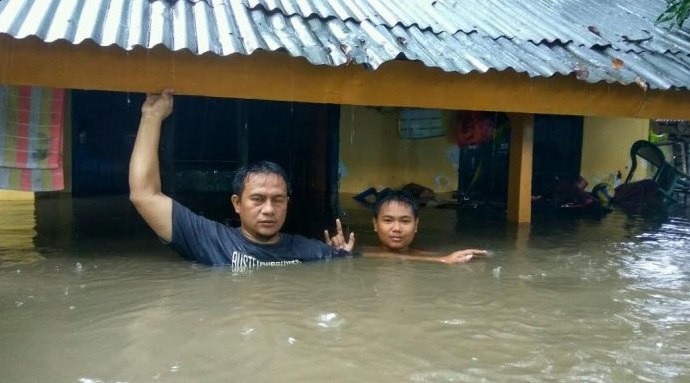}
\put(19,1){\color{white}\huge \textbf{Ground truth: 8}}
\end{overpic}}
& \rule{0pt}{1cm}
{\makecell{\dashuline{\textit{Regression}}\\ 6.2}}\\
\cline{3-3}
& & \rule{0pt}{1cm}
{\makecell{\dashuline{\textit{Regression++}}\\ 9.0}}\\
\cline{3-3}
& & \rule{0pt}{1cm}
{\makecell{\dashuline{\textit{Classification}~\cite{isprs-annals-IV-2-W5-5-2019}}\\ 6.7}}\\
\cline{3-3}
& & \rule{0pt}{1cm}
{\makecell{\dashuline{\textit{Reg+Rank}}\\ 8.3}}\\
\hline
\end{tabular}
\caption{Examples of test images and water level predictions for all four approaches \textit{Regression}, \textit{Regression++}, \textit{Classification}, and \textit{Reg+Rank}. Predicted water levels per image are written on the right side below each method, ground truth is given at the bottom of each image in white.}
\label{fig:quali_1}
\end{figure*}

\begin{figure*}

\centering\begin{tabular}{|c|c|c|}
\hline
\multirow{4}{*}{a} & \multirow{4}{*}[10pt]{\begin{overpic}[width=0.75\textwidth, height=5.2cm]{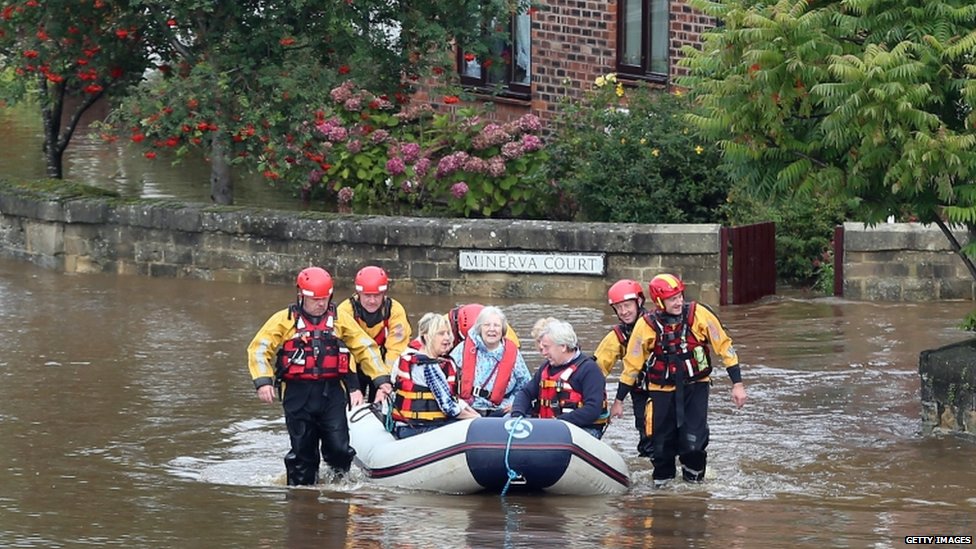}
\put(19,1){\color{white}\huge \textbf{Ground truth: 3}}
\end{overpic}}
& \rule{0pt}{0.75cm}
{\makecell{\dashuline{\textit{Regression}}\\ 5.7}}\\
\cline{3-3}
& & \rule{0pt}{0.75cm}
{\makecell{\dashuline{\textit{Regression++}}\\ 3.7}}\\
\cline{3-3}
& & \rule{0pt}{0.75cm}
{\makecell{\dashuline{\textit{Classification}~\cite{isprs-annals-IV-2-W5-5-2019}}\\ 3.3}}\\
\cline{3-3}
& & \rule{0pt}{0.75cm}
{\makecell{\dashuline{\textit{Reg+Rank}}\\ 3.1}}\\
\hline
\hline
\multirow{4}{*}{b} & \multirow{4}{*}[12pt]{\begin{overpic}[width=0.75\textwidth, height=6.0cm]{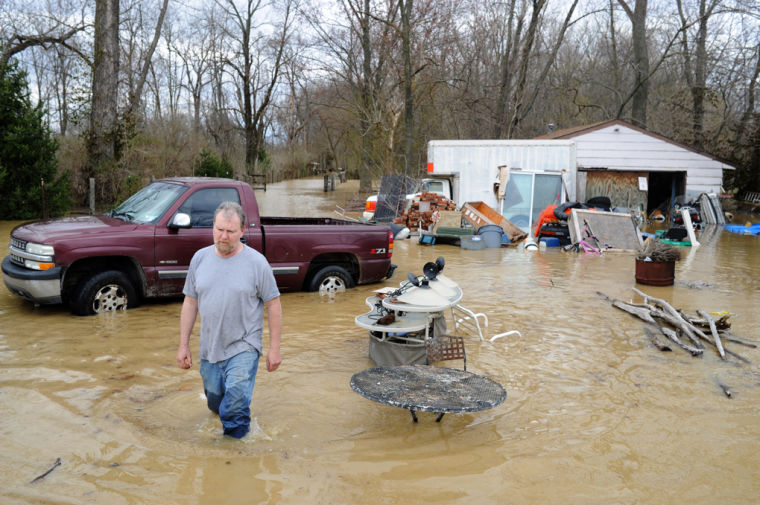}
\put(19,1){\color{white}\huge \textbf{Ground truth: 4}}
\end{overpic}}
& \rule{0pt}{1cm}
{\makecell{\dashuline{\textit{Regression}}\\ 3.5}}\\
\cline{3-3}
& & \rule{0pt}{1cm}
{\makecell{\dashuline{\textit{Regression++}}\\ 3.1}}\\
\cline{3-3}
& & \rule{0pt}{1cm}
{\makecell{\dashuline{\textit{Classification}~\cite{isprs-annals-IV-2-W5-5-2019}}\\ 3.0}}\\
\cline{3-3}
& & \rule{0pt}{1cm}
{\makecell{\dashuline{\textit{Reg+Rank}}\\ 3.0}}\\
\hline
\hline
\multirow{4}{*}{c} & \multirow{4}{*}[12pt]{\begin{overpic}[width=0.75\textwidth, height=6.0cm]{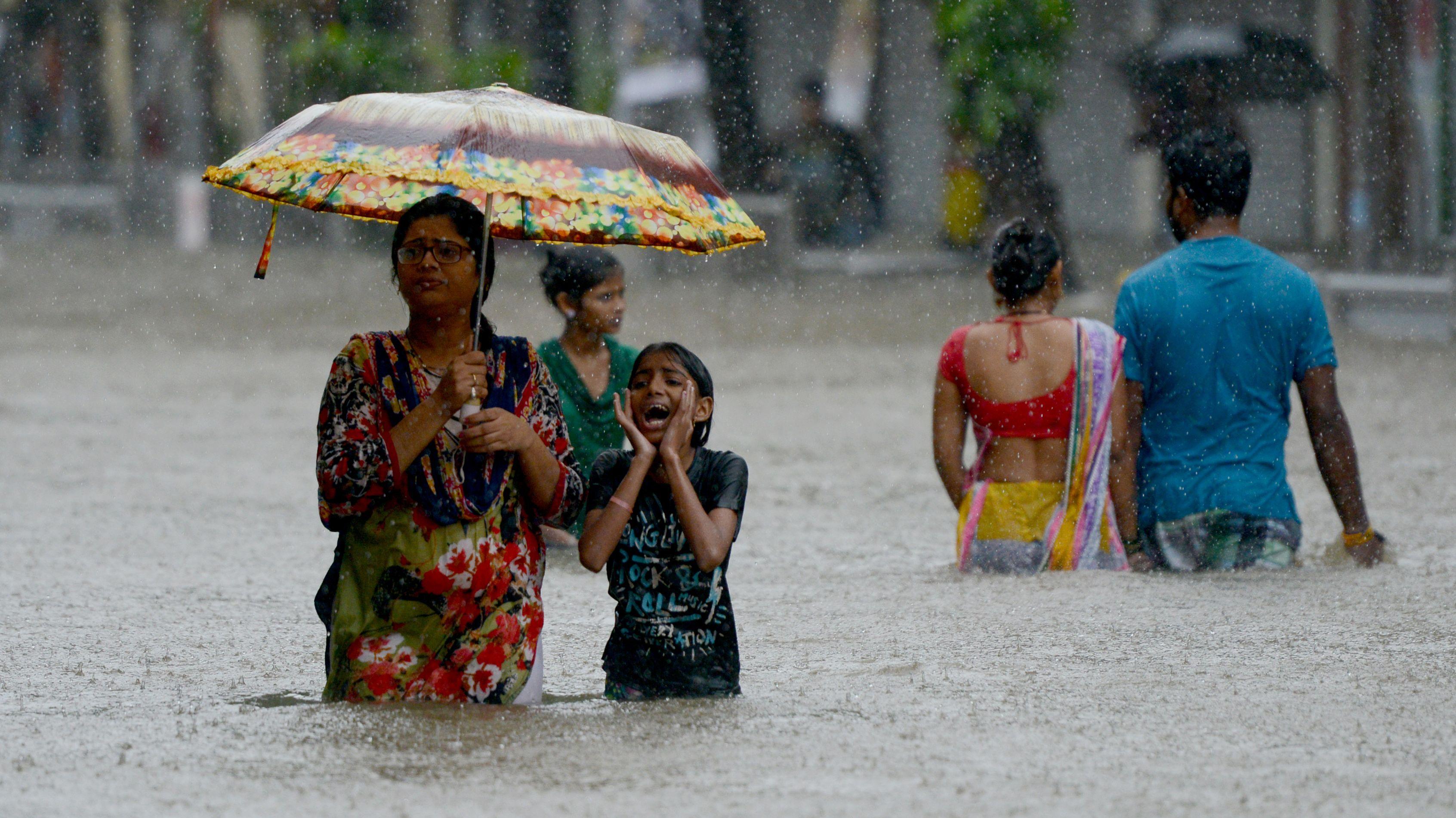}
\put(19,1){\color{white}\huge \textbf{Ground truth: 5}}
\end{overpic}}
& \rule{0pt}{1cm}
{\makecell{\dashuline{\textit{Regression}}\\ 5.1}}\\
\cline{3-3}
& & \rule{0pt}{1cm}
{\makecell{\dashuline{\textit{Regression++}}\\ 5.3}}\\
\cline{3-3}
& & \rule{0pt}{1cm}
{\makecell{\dashuline{\textit{Classification}~\cite{isprs-annals-IV-2-W5-5-2019}}\\ 4.8}}\\
\cline{3-3}
& & \rule{0pt}{1cm}
{\makecell{\dashuline{\textit{Reg+Rank}}\\ 4.7}}\\
\hline
\end{tabular}
\caption{Examples of test images and water level predictions for all four approaches \textit{Regression}, \textit{Regression++}, \textit{Classification}, and \textit{Reg+Rank}. Predicted water levels per image are written on the right side below each method, ground truth is given at the bottom of each image in white.}
\label{fig:quali_2}
\end{figure*}

We qualitatively illustrate water level predictions of all four tested models for some example test images of different cross-validation folds and water levels in Fig.~\ref{fig:quali_1} and Fig.~\ref{fig:quali_2}. 
As already indicated in Fig.~\ref{fig:box2} and Fig.~\ref{fig:box5}, all methods underestimate very high water levels (Fig.~\ref{fig:quali_1}a,b), which is most likely due to lack of sufficient training data. Those methods that have access to all available data of the \textsc{DeepFlood} dataset (\textit{Regression++} and \textit{Reg+Rank}), do perform better in several cases (Fig.~\ref{fig:quali_1}c).
Strong supervision via fine-grained, pixel-accurate object instance annotations (\textit{Classification}) improves training on a small dataset (\textsc{DF-Obj} subset) compared to weaker, per-image annotation (\textit{Regression}), as can be seen in Fig.~\ref{fig:quali_2}a. In this image, the true water level is somewhat hard to estimate for an automated method, as people at similar locations are in some cases upright in the water and in other cases seated in a boat. \textit{Regression} overestimates the water level by a large margin whereas \textit{Classification} is fairly accurate, presumably because with the help of explicit object labels it could learn how to handle people in boats, a situation of which there are several examples in the dataset. For more standard, rather frequent scenes with moderate flood levels 4 (Fig.~\ref{fig:quali_2}b) and 5 (Fig.~\ref{fig:quali_2}c) all methods work surprisingly well.  

\comment{
\begin{figure*}[th!]
    \centering
    \begin{subfigure}[t]{0.5\textwidth}
        \centering
        \includegraphics[width=1\linewidth]{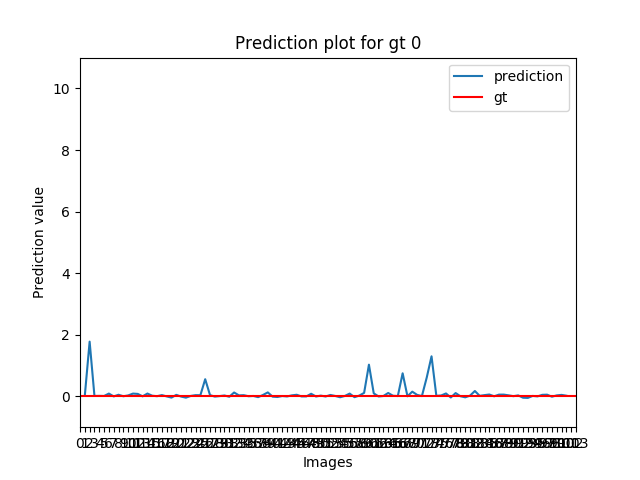}
        \caption{}
    \end{subfigure}%
    ~ 
    \begin{subfigure}[t]{0.5\textwidth}
        \centering
        \includegraphics[width=1\linewidth]{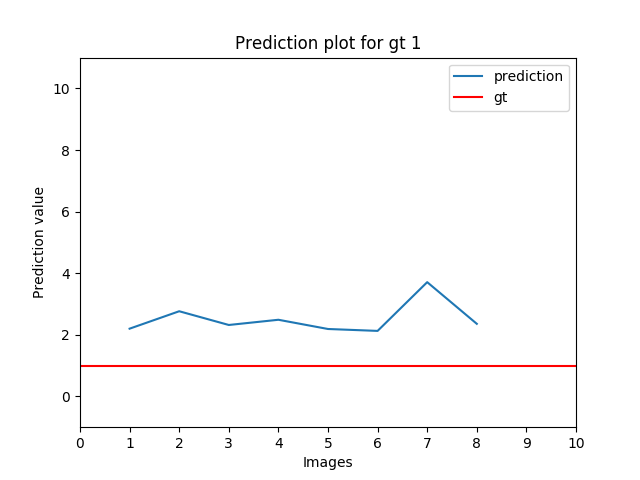}
        \caption{}
    \end{subfigure}
    \begin{subfigure}[t]{0.5\textwidth}
        \centering
        \includegraphics[width=1\linewidth]{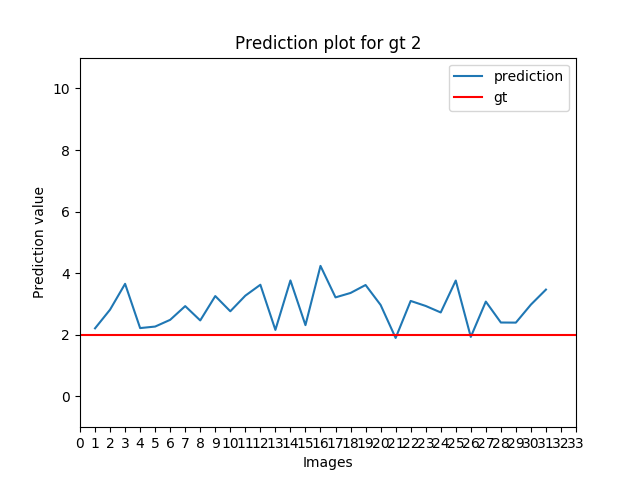}
        \caption{}
    \end{subfigure}%
    ~ 
    \begin{subfigure}[t]{0.5\textwidth}
        \centering
        \includegraphics[width=1\linewidth]{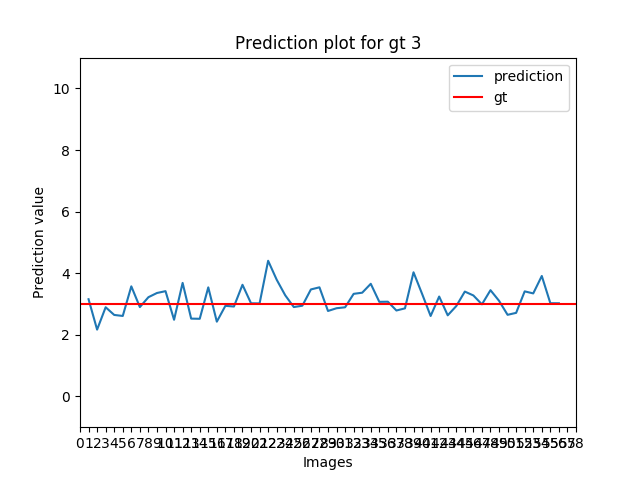}
        \caption{}
    \end{subfigure}
    \begin{subfigure}[t]{0.5\textwidth}
        \centering
        \includegraphics[width=1\linewidth]{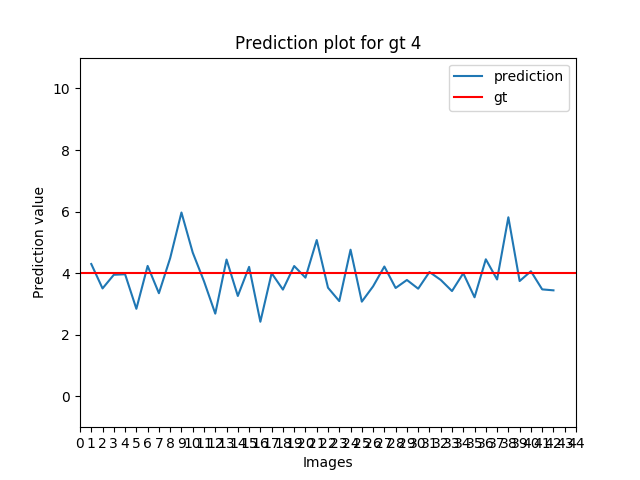}
        \caption{}
    \end{subfigure}%
    ~ 
    \begin{subfigure}[t]{0.5\textwidth}
        \centering
        \includegraphics[width=1\linewidth]{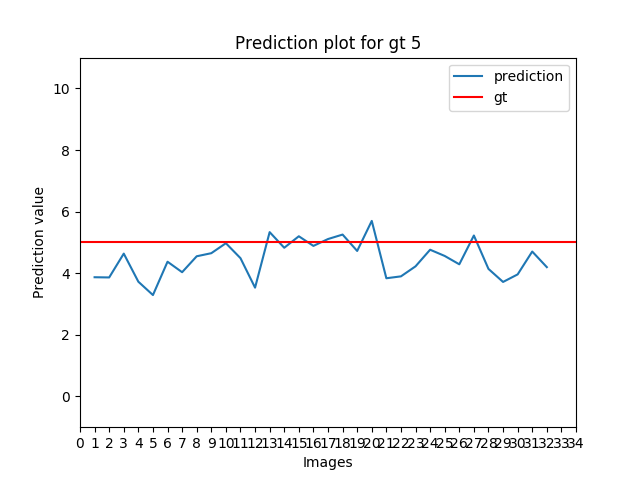}
        \caption{}
    \end{subfigure}
    \begin{subfigure}[t]{0.5\textwidth}
        \centering
        \includegraphics[width=1\linewidth]{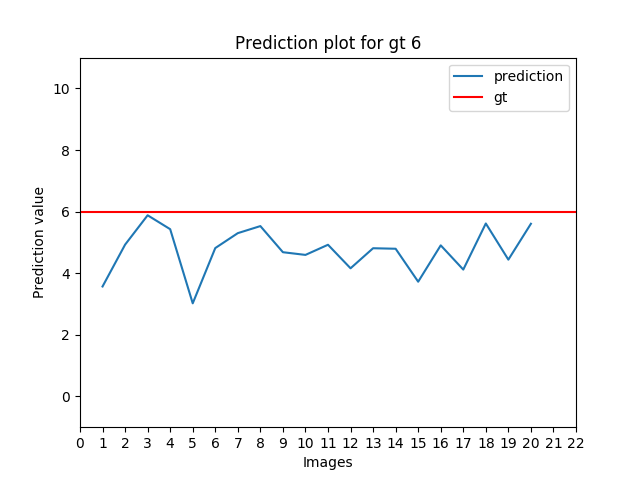}
        \caption{}
    \end{subfigure}%
    ~ 
    \begin{subfigure}[t]{0.5\textwidth}
        \centering
        \includegraphics[width=1\linewidth]{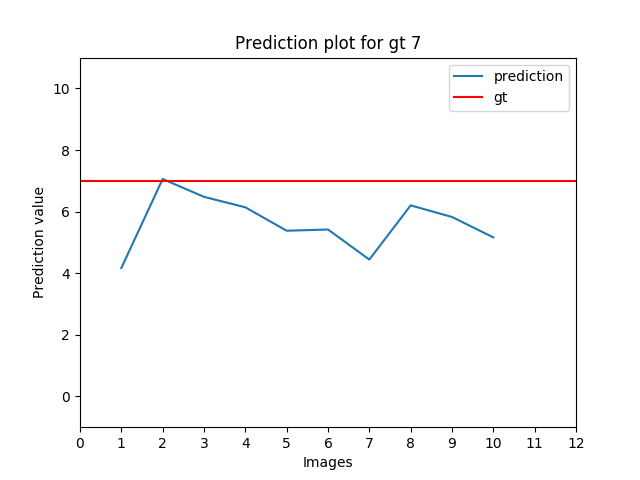}
        \caption{}
    \end{subfigure}
    \caption{Caption place holder}
    \label{fig:results}
\end{figure*}
\begin{figure*}[t]\ContinuedFloat
    \centering
    \begin{subfigure}[t]{0.5\textwidth}
        \centering
        \includegraphics[width=1\linewidth]{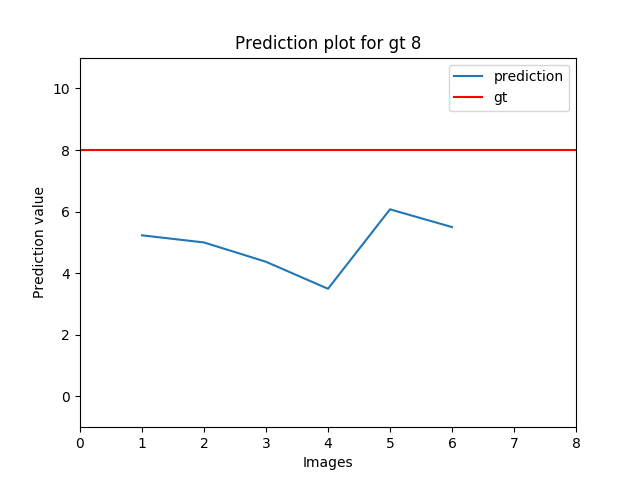}
        \caption{}
    \end{subfigure}%
    ~ 
    \begin{subfigure}[t]{0.5\textwidth}
        \includegraphics[width=1\linewidth]{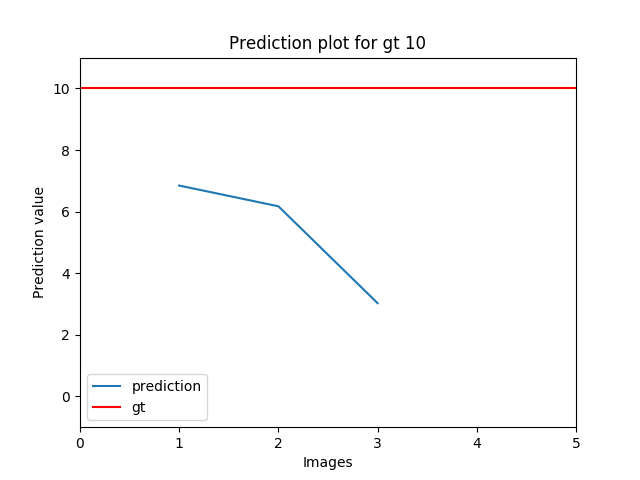}
        \caption{}
    \end{subfigure}%
    \caption{\textcolor{red}{Caption place holder}}
\end{figure*}
}


\begin{table}[t]
\centering
\begin{tabular}{|| c | c | c | c ||  c | c | c ||}
\hline\hline
\multirow{1}{4cm}{\textbf{Experiments}} & \multirow{1}{*}{\textbf{avgRMSE} [cm]} & \multirow{1}{*}{\textbf{stdDev} [cm]}  \\ 
\hline\hline
1M & 11.8 & 0.50\\
2M & 11.9 & 0.94\\ 
3M  & 11.6 &0.35\\ 
4M & 11.5 &0.19\\
6M & 11.3 & 0.64\\
\hline\hline
\end{tabular}
\caption{Ablation study for number of pairs. Average RMSE increases slowly with decreasing number of pairs. Results are in all cases computed with 5-fold cross-validation. The distinct folds have slightly different performances which results in minor \textbf{stdDev} value differences.}
\label{tab:ablation}
\end{table}
\subsection{Ablation Study}\label{sec:ablation}
In this section we analyse how the number of training image pairs affects the contribution of the ranking task. To that end we vary the number of distinct image pairs that are fed to the network during training from 1 million (1M) to 6 million (6M, \textit{Reg+Rank}). 
For this experiment, we subdivide the \textsc{DF-Img} subset into training and validation subsets. We use $5,005$ images to train the ranking task. Note that by using a mini-batch size of 5 this leads to $1,001$ iterations per epoch.

\begin{figure*}[htbp!]
\includegraphics[width=1\textwidth]{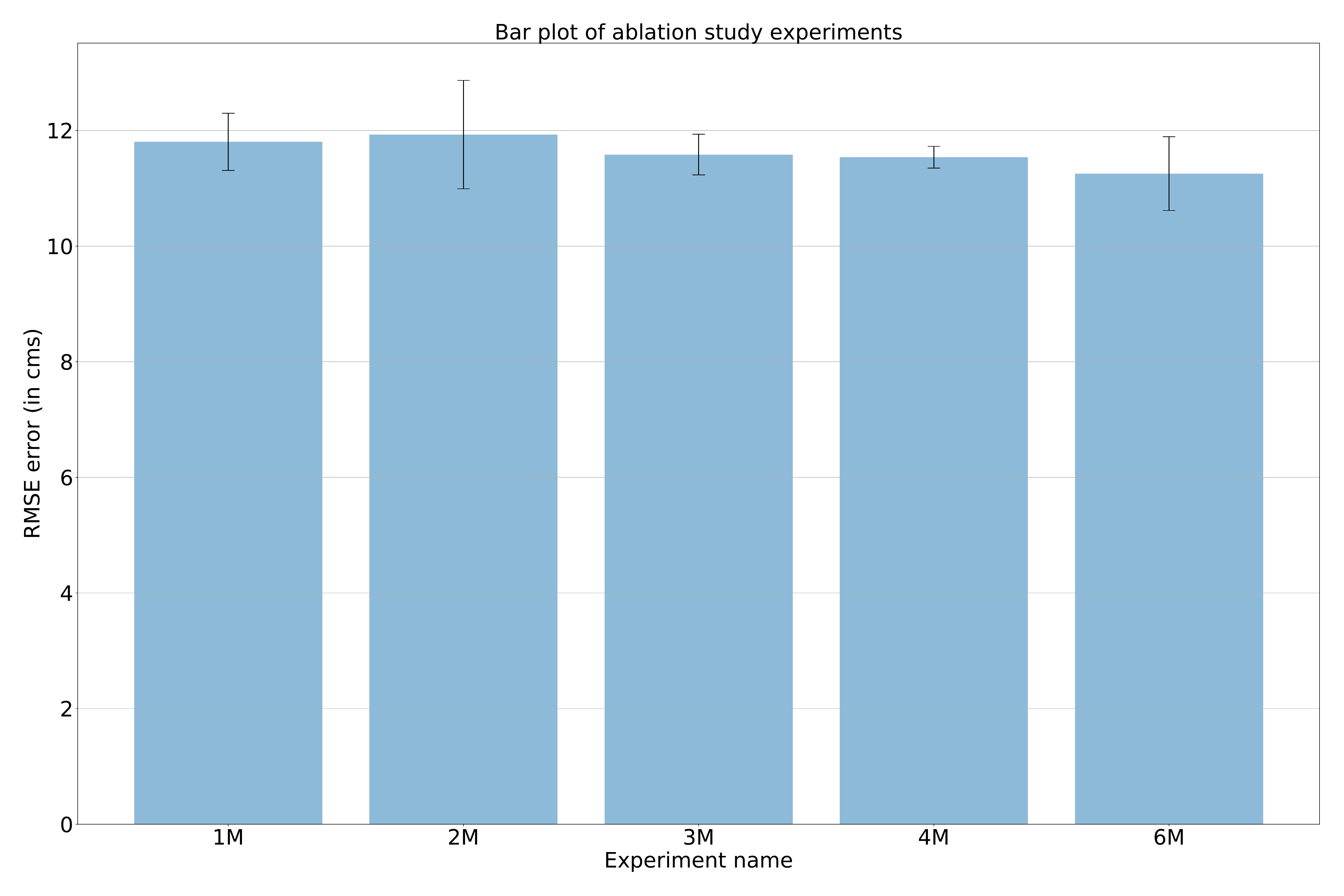}\par
\caption{Ablation study for number of pairs.}
\label{fig:ablation2}
\end{figure*}

Fig.~\ref{fig:ablation2} shows RMSE values (in centimeters) when increasing the maximum number of distinct image pairs from 1 million (1M) to six million (6M). As expected the RMSE decreases with increasing number of image pairs, but only slowly. 1M binary pair labels already bring a substantial improvement over the the regression baseline. We note that, while this may still seem like a high value, many of our automatically generated pair labels are redundant and could be derived from transitivity: whenever in a batch image $A$ has higher level than $B$ and $B$ has higher level than $C$, the pair $A$-$C$ need not be labeled.

\section{Conclusion}\label{sec:CONCLUSION}

We have proposed a fully automated method for water level estimation in social media images of flood events. The main idea of our approach is that it is much easier for a human annotator to decide in which of two images the water level is higher, rather than assign an absolute water level to a single image, let alone segment pixel-accurate object instance labels. We implement pairwise ranking as a form of weak supervision that regularises the training of the regressor.

The experimental comparison with a lower and upper performance bound for regression and an alternative classification scheme shows that the proposed weakly supervised method (\textit{Reg+Rank}) is able to perform almost as well as fully supervised regression with a much larger training set (\textit{Regression++}). Moreover, \textit{Reg+Rank} also outperforms \textit{Classification}~\cite{isprs-annals-IV-2-W5-5-2019}, although the necessary training data is, arguably, much easier to obtain. Weak supervision via pairwise ranking thus provides a promising alternative to costly and time-consuming, fine-grained labelling. 

We hope that our approach can help to overcome the label scarcity problem not only for water level prediction, but for many other regression tasks in the environmental and geo-sciences where collecting a sufficient amount of accurate labels is very laborious, and large datasets as needed for training deep learning are rare.


\bibliography{mybibfile}

\end{document}